\documentclass[runningheads]{llncs}

\usepackage{eccv}

\usepackage{eccvabbrv}

\usepackage{graphicx}
\usepackage{booktabs}

\usepackage{amsmath}
\usepackage{amssymb}
\usepackage{algorithm}
\usepackage{algpseudocode}

\usepackage{multirow}

\usepackage[accsupp]{axessibility}  

\usepackage{hyperref}

\usepackage{orcidlink}

\begin{document}

\title{Self-Routed Tensor Adapters for Parameter-Efficient Universal Visual Adaptation} 

\titlerunning{Self-Routed Tensor Adapters}

\author{Suraj Yadav}

\authorrunning{S.~Yadav}

\institute{
Indraprastha Institute of Information Technology Delhi \\
\email{suraj24098@iiitd.ac.in}
}

\maketitle

\begin{abstract}

Universal visual representations require adaptation mechanisms that adapt across heterogeneous domains without fragmenting knowledge into domain-specific modules. Parameter-efficient fine-tuning adapts frozen visual foundation models efficiently, but standard low-rank adapters use a fixed subspace for all inputs, which can be restrictive when domains differ in style, background, and semantic context. MoE-based adapters improve specialization through multiple expert pathways, but often rely on external routers and large expert banks, adding parameters and separating routing from adaptation.
We propose \textbf{Self-Routed Tensor Adapters}, a compact framework for multi-domain visual adaptation. SRTA projects each input into a low-rank space, computes routing weights from this representation using a learnable domain matrix, and uses these weights to blend slices of a shared Tucker core. This produces a sample-specific adaptation matrix without an external gating network, allowing shared visual factors to be reused while supporting domain-aware specialization. To strengthen pathway learning, we introduce a progressive depth-weighted routing objective that supervises routing decisions across adapter layers.
Across five heterogeneous multi-domain visual classification benchmarks, SRTA achieves competitive or slightly stronger average accuracy than MoE-style PEFT baselines while using substantially fewer trainable parameters. At rank 64, SRTA uses 2.77M parameters in the 4-domain setting compared with 9.52M for MoLoRA, and 3.00M in the 6-domain setting compared with 14.31M. Overall, SRTA offers an effective accuracy-parameter trade-off for adapting visual foundation models toward universal multi-domain representations.
\href{https://github.com/surajyadav-research/SRTA}{GitHub}
\end{abstract}

\section{Introduction}
\label{sec:intro}

Building universal visual representations \cite{dosovitskiy2020image, radford2021learning} requires models that remain effective across heterogeneous domains, tasks, and distribution shifts. In practice, however, adapting a large pre-trained vision backbone to multiple downstream domains remains challenging. Different domains may vary in visual style, background context, acquisition conditions, and object appearance. A useful adaptation mechanism should therefore preserve the reusable knowledge of the frozen foundation model while allowing compact, input-dependent specialization.

Parameter-efficient fine-tuning has emerged as a practical approach for adapting large pre-trained models while keeping the backbone frozen. LoRA \cite{hu2022lora} and related methods \cite{liu2024dora, kopiczko2024vera, jia2022visual} learn compact low-rank updates, but they usually apply the same adaptation subspace to every input. In heterogeneous multi-domain settings, this fixed subspace can be too restrictive: some domains require shared factors, while others require domain-specific adaptation. This may lead to negative transfer and uneven optimization across domains.

Recent work has addressed this limitation by combining PEFT with Mixture-of-Experts (MoE) architectures \cite{shazeer2017outrageously}. MoE-based adapters, such as MoLoRA \cite{zadouri2024pushing}, MOELoRA \cite{liu2024moe}, and MALoRA \cite{wang2025malora}, improve specialization by assigning inputs to multiple low-rank expert pathways. However, these methods typically compute routing probabilities using a separate gating network \cite{shazeer2017outrageously, zadouri2024pushing, liu2024moe}. This introduces additional routing-specific parameters and separates expert selection from the low-rank representation used for adaptation. As a result, routing decisions may become decoupled from the geometry of the learned adapter features, leading to redundant experts, unstable routing behavior, and inefficient sharing across related domains. Motivated by this limitation, we ask whether routing can instead be derived directly from the adapter representation itself.

In this paper, we study multi-domain visual adaptation from the perspective of universal representation learning. Instead of treating each domain as requiring an isolated expert, we ask whether heterogeneous domains can be organized within a shared, structured adaptation space. To this end, we propose \textbf{Self-Routed Tensor Adapters (SRTA)}, a parameter-efficient adaptation framework that formulates the update space as a Tucker-decomposed tensor \cite{tucker1966some, doi:10.1137/07070111X}. SRTA replaces discrete externally routed experts with a shared core tensor whose domain-indexed slices are dynamically blended for each input.

The key idea behind SRTA is \emph{intrinsic domain-aware routing}. Rather than introducing a separate gating network, SRTA derives routing probabilities directly from the interaction between the adapter's low-rank input projection and a learnable domain-coordinate matrix. This allows the adapter to route itself using the same representation that is used for adaptation. From a representation-learning perspective, the Tucker core serves as a shared adaptation basis, while the domain-coordinate matrix provides domain-aware coordinates over it. As a result, SRTA can reuse common visual factors across domains while softly specializing to domain-specific variations.

To encourage the learned pathways to remain both reusable and discriminative, we introduce a structural training objective. Specifically, we apply progressive depth-weighted routing supervision \cite{lee2015deeply} across adapter layers, providing direct learning signals to intermediate routing decisions and mitigating weak supervision in early layers. This objective helps reduce domain interference while preserving a compact shared adaptation space.

Our contributions are summarized as follows.

\begin{itemize}
    \item \textbf{Self-routed tensor adapters.}
    We introduce SRTA, a parameter-efficient adapter that computes routing probabilities directly from the adapter's low-rank representation, removing the need for a separate MoE gating network.

    \item \textbf{Input-conditioned Tucker adaptation.}
    We represent the adaptation space using a shared Tucker core tensor. For each input, SRTA dynamically blends core slices to produce a sample-specific low-rank adaptation matrix.

    \item \textbf{Progressive routing supervision.}
    We introduce a progressive depth weighted routing loss that improves domain-aware pathway learning across adapter layers.

    \item \textbf{Accuracy-parameter trade-off.}
    Across five multi-domain visual classification benchmarks, SRTA achieves competitive or slightly better average accuracy than MoE-style PEFT baselines while requiring substantially fewer parameters than MoLoRA.
\end{itemize}

\section{Related Work}

\subsection{Universal Visual Representations and Multi-Domain Adaptation}

A central goal of modern foundation models is to learn representations that remain transferable across tasks, domains, and distribution shifts. In visual recognition, this is closely related to domain generalisation, where models are trained on multiple source domains and expected to perform well under heterogeneous or unseen domain conditions. Benchmarks such as PACS~\cite{li2017deeper}, VLCS~\cite{torralba2011unbiased}, Office-Home~\cite{venkateswara2017deep}, Digits-DG~\cite{zhou2020deep}, and NICO++~\cite{zhang2023nico++} expose this challenge through variations in style, object appearance, background, acquisition conditions, and visual context. Although large pre-trained vision transformers provide strong general-purpose features~\cite{dosovitskiy2020image}, adapting them to multiple downstream domains remains non-trivial: full fine-tuning can overwrite useful pre-trained knowledge, while a single shared adaptation module may be insufficient for domains with conflicting visual statistics. This motivates adaptation mechanisms that preserve reusable representations while allowing controlled domain-specific specialization. Our work follows this direction by introducing a self-routed Tucker-decomposed adaptation space that blends shared and domain-aware factors.

\subsection{Parameter-Efficient Fine-Tuning}

Parameter-Efficient Fine-Tuning (PEFT) adapts large pre-trained models by updating only a small subset of parameters while keeping the backbone frozen. Prompt-based methods, such as Prefix-Tuning~\cite{li2021prefix} and Prompt-Tuning~\cite{lester2021power}, optimize continuous prompts, while Low-Rank Adaptation (LoRA)~\cite{hu2022lora} learns trainable low-rank updates to frozen weight matrices and has become a widely used PEFT technique. Several extensions improve efficiency or adaptation quality: DoRA~\cite{liu2024dora} decomposes updates into magnitude and direction components, while VeRA~\cite{kopiczko2024vera} and VB-LoRA~\cite{li2024vb} reduce trainable parameters through shared or reused adaptation components. However, most PEFT methods rely on static adaptation spaces, which can be limiting in multi-domain settings where a single low-rank update may suffer from negative transfer and domain interference. In contrast, our method introduces an input-dependent self-routing mechanism that dynamically blends adaptation pathways while remaining parameter-efficient.

\subsection{Mixture-of-Experts for Multi-Domain Adaptation}

Mixture-of-Experts (MoE) \cite{fedus2022switch} architectures improve multi-domain adaptation by assigning inputs to specialized expert modules. In PEFT, MoE-based adapters combine low-rank efficiency with expert specialization: MoLoRA~\cite{zadouri2024pushing} extends LoRA with multiple low-rank experts and a routing mechanism, MOELoRA~\cite{liu2024moe} introduces task-conditioned gating to combine LoRA experts for multi-task adaptation, and MALoRA~\cite{wang2025malora} improves parameter efficiency by sharing parts of the low-rank structure across experts. Despite their effectiveness, these methods commonly rely on external gating networks to compute routing probabilities, creating a separation between the adapter representation space and the routing mechanism. This decoupling can add parameters, increase computation, and lead to redundant expert assignments. Our work removes the external router and derives routing probabilities directly from the low-rank adapter representation through its interaction with a learnable domain-coordinate matrix, aligning expert blending with the representation geometry used for adaptation.

\subsection{Tensor Decompositions and Factorized Adapters}

Tensor decompositions provide a principled way to represent high-dimensional parameter spaces using compact multilinear factors. Classical decompositions such as CP~\cite{carroll1970analysis} and Tucker decomposition~\cite{tucker1966some, doi:10.1137/07070111X} have been widely used for compression, factor analysis, and efficient parameterization. In neural adaptation, tensor and factorized methods have also been explored to reduce fine-tuning cost; for example, FacT~\cite{jie2023fact} applies tensor-based factorization to lightweight adaptation in vision transformers. Most tensor-based adaptation methods use decomposition primarily as a compression mechanism. In contrast, our work uses a Tucker-decomposed parameter space as a dynamic routing structure, where the shared core tensor captures reusable adaptation factors and the domain-coordinate matrix defines domain-aware coordinates for blending them. This enables both shared and domain-specific structure to be represented within a single compact adaptation module.

\section{Methodology}

\subsection{Problem Setup}

We consider multi-domain parameter-efficient adaptation of a frozen pre-trained vision model. Let 
$W_0 \in \mathbb{R}^{d_{\mathrm{in}} \times d_{\mathrm{out}}}$ denote a frozen weight matrix in a transformer layer, and let 
$x \in \mathbb{R}^{B \times T \times d_{\mathrm{in}}}$ be an input sequence with batch size $B$ and sequence length $T$. 
During training, each sample is associated with a class label $y^{*}$ and a domain label $d^{*}$. The class label supervises the main recognition objective, while the domain label is used only as auxiliary supervision for routing. At inference time, SRTA does not require the domain label; routing probabilities are computed directly from the input representation.

The goal is to learn a compact input-conditioned adaptation module that supports heterogeneous domains while preserving a shared representation space. In standard Low-Rank Adaptation (LoRA), the update is parameterized as

\begin{equation}
    \Delta W = AB,
\end{equation}

where $A \in \mathbb{R}^{d_{\mathrm{in}} \times r}$ and $B \in \mathbb{R}^{r \times d_{\mathrm{out}}}$ are trainable low-rank matrices. The adapted layer output is then

\begin{equation}
    y = xW_0 + s(xAB),
\end{equation}

where $s$ is a scaling factor. While this static low-rank update is efficient, it uses the same adaptation subspace for all inputs. This can be restrictive in heterogeneous multi-domain settings, where different domains may require partially shared but also domain-specific adaptation pathways.

MoE-based LoRA methods address this limitation by introducing $N$ low-rank experts:

\begin{equation}
    y = xW_0 + s \sum_{t=1}^{N} \alpha_t(x A_t B_t),
\end{equation}

where $\alpha_t$ denotes the routing probability for expert $t$. These routing probabilities are usually produced by an external gating network. Although this enables input-dependent specialization, the router is separated from the adapter representation itself. This introduces additional routing-specific parameters and may decouple expert selection from the low-rank representation being adapted.

\subsection{Self-Routed Tensor Adapters}

To avoid external routing while still enabling domain-aware specialization, we propose \textbf{Self-Routed Tensor Adapters (SRTA)}. The key idea is to formulate the adaptation space as a Tucker-decomposed tensor and derive routing probabilities intrinsically from the adapter's own low-rank representation.

Instead of representing the update using a single low-rank matrix product, SRTA introduces four trainable components:

\begin{itemize}
    \item an input projection matrix $V \in \mathbb{R}^{d_{\mathrm{in}} \times r_1}$,
    \item an output projection matrix $U \in \mathbb{R}^{r_2 \times d_{\mathrm{out}}}$,
    \item a shared Tucker core tensor $\mathcal{G} \in \mathbb{R}^{r_3 \times r_1 \times r_2}$,
    \item a domain-coordinate matrix $C \in \mathbb{R}^{r_1 \times r_3}$.
\end{itemize}

Here, $r_1$ and $r_2$ are the low-rank input and output dimensions, while $r_3$ denotes the number of domain-aware adaptation pathways. In our experiments, $r_3$ is set to the number of domains in the benchmark. The columns of $C$ act as learnable pathway coordinates in the low-rank adapter space.

\begin{figure}[H]
    \centering
    \includegraphics[width=\textwidth]{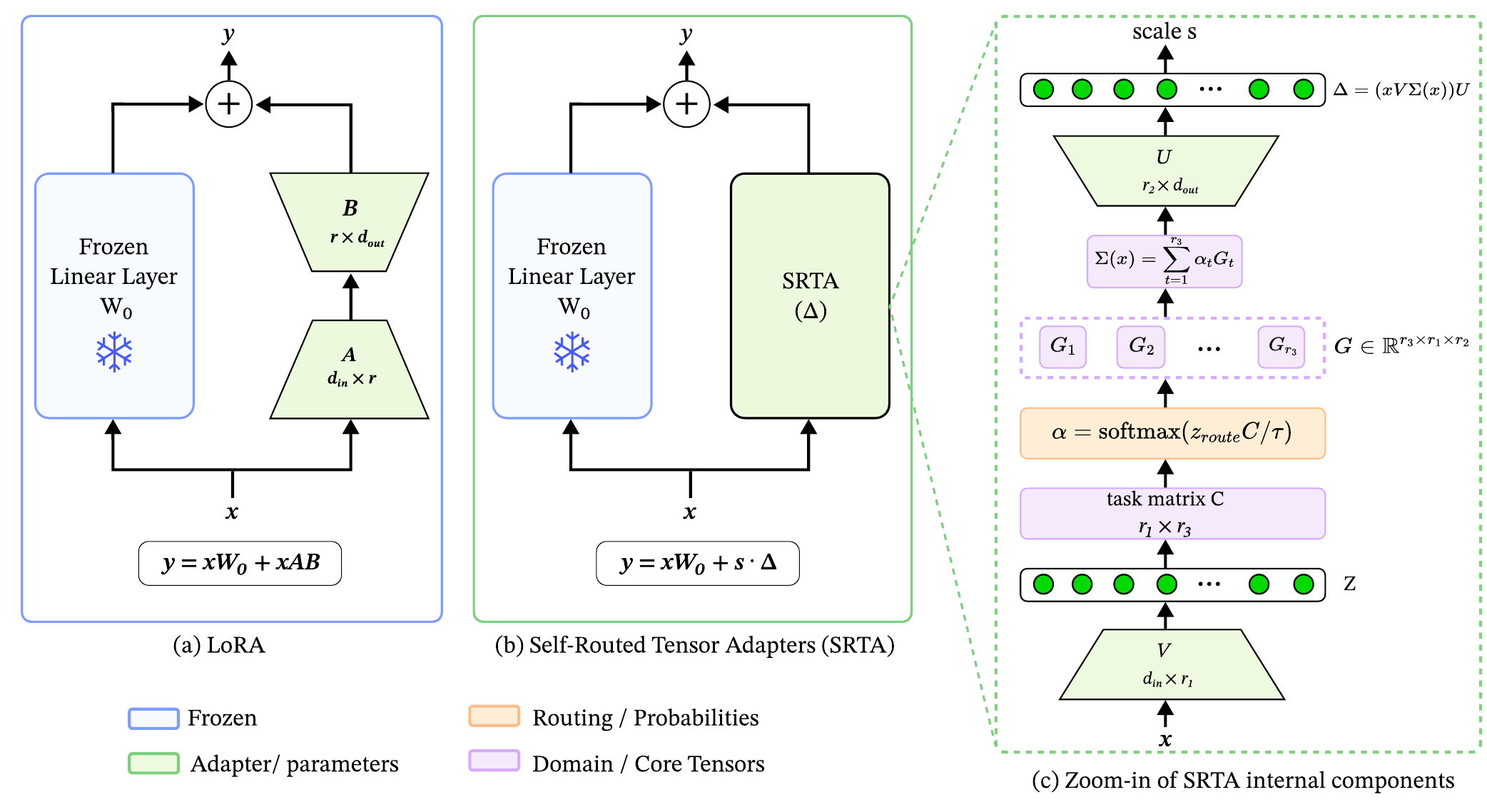}
    \caption{\textbf{Overview of the Self-Routed Tensor Adapter pipeline}. The input projection $V$ and domain-coordinate matrix $C$ generate intrinsic routing probabilities, which are used to blend slices of the shared Tucker core tensor $\mathcal{G}$. This removes the need for an external MoE gating network.}
    \label{fig:srta_pipeline}
\end{figure}

From a representation-learning perspective, the shared core tensor $\mathcal{G}$ stores reusable adaptation factors, while the domain-coordinate matrix $C$ provides coordinates for routing over this shared basis. This allows SRTA to softly combine shared and domain-specific factors instead of allocating each input to a fully isolated expert.

\subsection{Intrinsic Self-Routing}

Given an input sequence $x$, SRTA first projects it into the low-rank adapter space:

\begin{equation}
    z = xV,
    \qquad
    z \in \mathbb{R}^{B \times T \times r_1}.
\end{equation}

Instead of relying only on the global classification token, we compute a routing signature by averaging the projected patch-token representations. Using zero-based indexing, we treat token 0 as the $\mathrm{[CLS]}$ token, we omit this token and aggregate over the remaining patch tokens:

\begin{equation}
    z_{\mathrm{route}} =
    \frac{1}{T - 1}
    \sum_{i=1}^{T-1} z[:, i, :],
    \qquad
    z_{\mathrm{route}} \in \mathbb{R}^{B \times r_1}.
\end{equation}

This routing signature encourages routing decisions to depend on distributed patch-level visual evidence rather than a single global token representation.

Routing logits are then computed by comparing the pooled low-rank representation with the learnable domain-coordinate matrix:

\begin{equation}
    q =
    \frac{z_{\mathrm{route}} C}{\tau},
    \qquad
    q \in \mathbb{R}^{B \times r_3},
\end{equation}

where $\tau$ is a temperature hyperparameter that controls the sharpness of the routing distribution. The routing probabilities are obtained using a softmax function:

\begin{equation}
    \alpha =
    \mathrm{Softmax}(q),
    \qquad
    \alpha \in \mathbb{R}^{B \times r_3}.
\end{equation}

Unlike MoE-based adapters, SRTA does not introduce a separate gating network. The routing probabilities are computed from the same low-rank representation that is used for adaptation. Therefore, routing and feature adaptation are coupled within the same representation space.

\subsection{Dynamic Tucker Core Blending}

The routing probabilities $\alpha$ are used to dynamically blend the domain-pathway slices of the shared Tucker core tensor. Let 
$\mathcal{G}_t \in \mathbb{R}^{r_1 \times r_2}$ denote the $t$-th slice of the core tensor. For each input sample $b$, SRTA constructs a sample-specific core matrix:

\begin{equation}
    \Sigma_b(x)
    =
    \sum_{t=1}^{r_3}
    \alpha_{b,t} \mathcal{G}_t,
    \qquad
    \Sigma_b(x) \in \mathbb{R}^{r_1 \times r_2}.
\end{equation}

The adapter output for sample $b$ is then computed as

\begin{equation}
    y_{\mathrm{adapt},b}
    =
    (z_b \Sigma_b(x))U,
\end{equation}

where $z_b \in \mathbb{R}^{T \times r_1}$ is the low-rank projected representation of sample $b$. The final adapted layer output is

\begin{equation}
    y_b
    =
    x_b W_0
    +
    s \cdot y_{\mathrm{adapt},b},
\end{equation}

where $s$ is the adapter scaling factor.

This formulation can be viewed as an input-conditioned Tucker factorization of the adapter update. Instead of selecting from independent expert matrices, SRTA learns a shared tensor basis and constructs a sample-specific adaptation matrix by softly blending core slices. This allows related domains to share adaptation factors while still permitting domain-specific specialization.

\subsection{Progressive Depth-Weighted Routing Supervision}

Routing decisions are computed independently at each adapter layer. However, if routing is supervised only through the final classification loss, the routing parameters in earlier layers may receive weak or delayed learning signals. To address this, we introduce progressive depth-weighted routing supervision.

Let $M$ denote the total number of adapter layers. For adapter layer $\ell$, let $q^{(\ell)}$ be the routing logits and let $d^{*}$ be the ground-truth domain label. We define the layer-wise routing loss as

\begin{equation}
    \mathcal{L}_{\mathrm{route}}^{(\ell)}
    =
    \mathcal{L}_{\mathrm{CE}}
    \left(
    q^{(\ell)}, d^{*}
    \right),
\end{equation}

where $\mathcal{L}_{\mathrm{CE}}$ is the cross-entropy loss. Each layer is assigned a progressive depth weight:

\begin{equation}
    w_\ell =
    \frac{\ell}{M}.
\end{equation}

This assigns smaller weights to shallow layers and larger weights to deeper layers, reflecting the intuition that deeper representations are more semantically abstract and more suitable for domain-discriminative routing. The weighted routing supervision provides direct learning signals to intermediate routing decisions while keeping inference unchanged.

\subsection{Training Objective}

The final training objective combines the main classification loss with the auxiliary depth-weighted routing loss:

\begin{equation}
    \mathcal{L}_{\mathrm{total}}
    =
    \mathcal{L}_{\mathrm{CE}}
    \left(
    \hat{y}, y^{*}
    \right)
    +
    \lambda_{\mathrm{route}}
    \frac{1}{M}
    \sum_{\ell=1}^{M}
    w_\ell
    \mathcal{L}_{\mathrm{route}}^{(\ell)}.
\end{equation}

Here, $\hat{y}$ is the model prediction, $y^{*}$ is the ground-truth class label, $d^{*}$ is the domain label used inside the routing loss, and $\lambda_{\mathrm{route}}$ controls the strength of the auxiliary routing supervision.

The domain label is used only during training for the progressive depth-weighted routing loss. During inference, SRTA computes routing probabilities directly from the input representation and does not require domain labels. Therefore, the auxiliary routing supervision improves pathway learning without adding inference-time inputs or parameters.

\section{Experiments}

\subsection{Experimental Setup}


\paragraph{Datasets.}
We conduct experiments on five widely used multi-domain visual benchmarks: PACS~\cite{li2017deeper}, VLCS~\cite{torralba2011unbiased}, Office-Home~\cite{venkateswara2017deep}, Digits-DG~\cite{zhou2020deep}, and NICO++~\cite{zhang2023nico++}. These datasets contain heterogeneous domains with variations in visual style, object appearance, background context and semantic overlap. For each dataset, we treat every domain as a distinct task or domain pathway, allowing us to evaluate whether SRTA can learn reusable adaptation factors while maintaining domain-aware routing behavior. PACS, VLCS, Office-Home, and Digits-DG contain four domains, while NICO++ contains six domains in our experimental setup.

\paragraph{Data Split.}
For each benchmark, we use an 80/20 train-test split. From the training portion, we uniformly sample 2400 images, using 2000 images for training and 400 images for validation. This controlled setup allows us to evaluate adaptation under limited data while maintaining balanced domain supervision.

\paragraph{Backbone.}
All methods use a frozen \texttt{vit-base-patch16-224-in21k}~\cite{dosovitskiy2020image} backbone. Only the adaptation modules and the classification head are trained. This setting isolates the effect of each PEFT method and evaluates its ability to adapt a shared pre-trained representation to heterogeneous visual domains.

\paragraph{Baselines.}
We compare SRTA against representative PEFT and MoE-PEFT baselines:

\begin{itemize}
    \item \textbf{LoRA}~\cite{hu2022lora}: standard low-rank adaptation.
    \item \textbf{DoRA}~\cite{liu2024dora}: weight-decomposed low-rank adaptation.
    \item \textbf{MOELoRA}~\cite{liu2024moe}: MOE-LoRA with task-conditioned expert routing.
    \item \textbf{MALoRA}~\cite{wang2025malora}: MoE-LoRA with shared low-rank structure.
    \item \textbf{MoLoRA}~\cite{zadouri2024pushing}: MoE-based LoRA with multiple low-rank experts.
\end{itemize}

\paragraph{Implementation Details.}
Adapters are injected into the Query and Value projection matrices of all transformer layers unless otherwise specified. All models are trained for 30 epochs using the AdamW optimizer with a batch size of 64. The learning rate is set to $5 \times 10^{-4}$ with a weight decay of $0.01$ and a warmup ratio of $0.1$.

For SRTA, the routing temperature is set to $\tau = 0.1$, and the adapter scaling factor is set to $s = 4.0$. The routing loss coefficient is set to $\lambda_{\mathrm{route}} = 1$. Unless otherwise stated, the Tucker ranks are set to $r_1 = 64$ and $r_2 = 64$, while $r_3$ is set to the number of domains in the dataset.

\subsection{Main Results}
Table~\ref{tab:main_results} reports the classification accuracy across five heterogeneous multi-domain visual benchmarks at rank 64. SRTA achieves the highest average accuracy of $88.1\%$, slightly outperforming MoLoRA. While the absolute accuracy gain is modest, SRTA achieves this result without an external gating network, suggesting that its main advantage lies in the accuracy-parameter trade-off rather than a large standalone accuracy improvement.

\begin{table}[t]
\centering
\caption{Accuracy (\%) across heterogeneous multi-domain visual benchmarks (Rank 64).}
\label{tab:main_results}
\begin{tabular}{lcccccc}
\toprule
\textbf{Method} & \textbf{PACS} & \textbf{VLCS} & \textbf{Office-Home} & \textbf{Digits-DG} & \textbf{NICO++} & \textbf{Average} \\
\midrule
LoRA    & $\underline{95.1}{\scriptstyle \pm 0.1}$ & $82.8{\scriptstyle \pm 0.7}$ & $77.3{\scriptstyle \pm 0.3}$ & $\mathbf{93.8}{\scriptstyle \pm 0.3}$ & $77.6{\scriptstyle \pm 0.3}$ & $85.3$ \\
DoRA    & $95.0{\scriptstyle \pm 0.2}$ & $82.8{\scriptstyle \pm 0.3}$ & $77.3{\scriptstyle \pm 0.3}$ & $\mathbf{93.8}{\scriptstyle \pm 0.4}$ & $77.4{\scriptstyle \pm 0.3}$ & $85.3$ \\
MALoRA  & $94.3{\scriptstyle \pm 0.5}$ & $82.8{\scriptstyle \pm 0.1}$ & $85.0{\scriptstyle \pm 0.0}$ & $90.2{\scriptstyle \pm 0.4}$ & $82.3{\scriptstyle \pm 0.3}$ & $86.9$ \\
MOELoRA & $94.5{\scriptstyle \pm 0.3}$ & $84.2{\scriptstyle \pm 0.5}$ & $\underline{85.2}{\scriptstyle \pm 0.0}$ & $88.9{\scriptstyle \pm 1.3}$ & $\mathbf{83.3}{\scriptstyle \pm 1.0}$ & $87.2$ \\
MoLoRA  & $\underline{95.1}{\scriptstyle \pm 0.3}$ & $\underline{84.4}{\scriptstyle \pm 0.9}$ & $\mathbf{85.3}{\scriptstyle \pm 0.6}$ & $91.7{\scriptstyle \pm 0.5}$ & $\mathbf{83.3}{\scriptstyle \pm 0.3}$ & $\underline{88.0}$ \\
\textbf{SRTA} & $\mathbf{95.2}{\scriptstyle \pm 0.7}$ & $\mathbf{84.7}{\scriptstyle \pm 0.3}$ & $84.8{\scriptstyle \pm 0.7}$ & $\underline{92.7}{\scriptstyle \pm 0.6}$ & $\underline{83.0}{\scriptstyle \pm 0.4}$ & $\mathbf{88.1}$ \\
\bottomrule
\end{tabular}
\end{table}

The results show that static PEFT methods, such as LoRA and DoRA, perform competitively on PACS and Digits-DG, but are weaker on more heterogeneous benchmarks such as Office-Home and NICO++. This suggests that a single static low-rank adaptation subspace may be insufficient when domains exhibit diverse visual styles, backgrounds, and semantic contexts. MoE-based methods improve performance on these harder benchmarks by introducing multiple expert pathways, with MoLoRA and MOELoRA performing strongly on Office-Home and NICO++.

SRTA remains competitive with these MoE-based methods while using intrinsic self-routing and a shared Tucker core instead of an external router or independent expert banks. It achieves the best results on PACS and VLCS, and strong performance on Digits-DG and NICO++. Although MoLoRA performs slightly better on Office-Home, SRTA achieves the best overall average accuracy while maintaining a more compact adaptation structure. These results indicate that input-conditioned Tucker-based routing can effectively organize heterogeneous domains within a shared adaptation space.

\subsection{Parameter Efficiency}

A key advantage of SRTA is that it avoids the parameter overhead introduced by externally routed MoE adapters. Table~\ref{tab:param_efficiency_ranks} compares the number of trainable parameters across different rank settings for both 4-domain and 6-domain configurations, excluding the classifier to reflect only the cost of the adaptation modules. At rank 64, SRTA requires 2.77M trainable parameters in the 4-domain setting and 3.00M parameters in the 6-domain setting, whereas MoLoRA requires 9.52M and 14.31M parameters, respectively. This corresponds to about $3.4\times$ fewer parameters than MoLoRA in the 4-domain setting and about $4.8\times$ fewer parameters in the 6-domain setting.

These results show that SRTA's main advantage is not simply accuracy, but achieving MoE-level adaptation performance with a substantially smaller parameter budget. Compared with LoRA and DoRA, SRTA introduces only a modest parameter increase while enabling input-dependent domain-aware routing. Overall, these results show that strong multi-domain adaptation does not require large independent expert banks; instead, a compact Tucker-decomposed shared tensor space can provide an effective accuracy-parameter trade-off by reusing common adaptation factors while supporting domain-specific specialization.

\begin{table}[t]
\centering
\caption{Parameter comparison across ranks for 4-domain and 6-domain settings. Parameter counts exclude the classifier.}
\label{tab:param_efficiency_ranks}
\small
\setlength{\tabcolsep}{4pt} 
\begin{tabular}{lcccccccc}
\toprule
\multirow{2}{*}{Method}
& \multicolumn{4}{c}{4 Domains}
& \multicolumn{4}{c}{6 Domains} \\
\cmidrule(lr){2-5} \cmidrule(lr){6-9}
& $r$=8 & $r$=16 & $r$=32 & $r$=64
& $r$=8 & $r$=16 & $r$=32 & $r$=64 \\
\midrule
LoRA    & 0.29M & 0.59M & 1.18M & 2.36M & 0.29M & 0.59M & 1.18M & 2.36M \\
DoRA    & 0.31M & 0.61M & 1.20M & 2.38M & 0.31M & 0.61M & 1.20M & 2.38M \\
MALoRA  & 0.53M & 0.98M & 1.90M & 3.82M & 0.75M & 1.35M & 2.59M & 5.17M \\
MOELoRA & 0.30M & 0.59M & 1.18M & 2.36M & 0.44M & 0.89M & 1.77M & 3.54M \\
MoLoRA  & 1.26M & 2.44M & 4.80M & 9.52M & 1.93M & 3.70M & 7.23M & 14.31M \\
SRTA    & 0.31M & 0.62M & 1.29M & 2.77M & 0.35M & 0.68M & 1.38M & 3.00M \\
\bottomrule
\end{tabular}
\end{table}

\subsection{Ablation on Loss Components}

We evaluate the effect of the progressive depth-weighted routing supervision by varying the auxiliary routing loss coefficient $\lambda_{\mathrm{route}}$. Table~\ref{tab:loss_ablation} compares SRTA without routing supervision $(\lambda_{\mathrm{route}}=0.0)$ and with routing supervision $(\lambda_{\mathrm{route}}=1.0)$ at rank 64. Adding the routing loss improves the average accuracy from $87.1\%$ to $88.1\%$, showing that direct supervision of intermediate routing decisions helps the adapter learn more effective domain-aware pathways. The improvement is especially clear on VLCS, Office-Home, and NICO++, while PACS remains unchanged, suggesting that routing supervision is most useful when domains are more heterogeneous or semantically overlapping. These results confirm that progressive depth-weighted routing supervision strengthens domain-aware pathway learning and improves the overall stability of SRTA without adding inference-time parameters.

\begin{table}[t]
\centering
\caption{Ablation of SRTA loss components at rank $64$.}
\label{tab:loss_ablation}
\begin{tabular}{ccccccc}
\toprule
$\lambda_{\mathrm{route}}$ & \textbf{PACS} & \textbf{VLCS} & \textbf{Office-Home} & \textbf{Digits-DG} & \textbf{NICO++} & \textbf{Average} \\
\midrule
0.0 & $\mathbf{95.2}{\scriptstyle \pm 0.3}$ & $82.3{\scriptstyle \pm 0.9}$ & $83.5{\scriptstyle \pm 0.3}$ & $92.6{\scriptstyle \pm 0.4}$ & $81.7{\scriptstyle \pm 0.6}$ & $87.1$ \\
1.0 & $\mathbf{95.2}{\scriptstyle \pm 0.7}$ & $\mathbf{84.7}{\scriptstyle \pm 0.4}$ & $\mathbf{84.8}{\scriptstyle \pm 0.8}$ & $\mathbf{92.7}{\scriptstyle \pm 0.7}$ & $\mathbf{83.0}{\scriptstyle \pm 0.4}$ & $\mathbf{88.1}$ \\
\bottomrule
\end{tabular}
\end{table}

\subsection{Ablation on Adapter Ranks}

To analyze the trade-off between adaptation capacity and parameter efficiency, we evaluate all PEFT and MoE-PEFT methods under different rank configurations. For LoRA and DoRA, we vary the low-rank dimension $r$. For MALoRA, we vary the shared rank and expert rank. For MoELoRA, we vary the number of routed low-rank expert dimensions following its baseline configuration. For MoLoRA, we vary the expert rank. For SRTA, we vary the Tucker ranks $r_1$ and $r_2$, while keeping $r_3$ equal to the number of domains. Table~\ref{tab:rank_ablation_all} reports the accuracy across all benchmarks.

Across static PEFT methods, increasing the rank generally improves performance, but the gains saturate at higher ranks. LoRA and DoRA show steady improvements from rank 8 to rank 64, but they remain weaker on more heterogeneous datasets such as Office-Home and NICO++, where a single static low-rank subspace is less effective. MoE-based methods perform strongly by introducing multiple expert pathways, but their improvements are not always monotonic with rank. For example, MoLoRA achieves highly competitive results at rank 32, but its performance does not consistently improve at rank 64, suggesting that simply increasing expert capacity does not necessarily resolve multi-domain interference.

SRTA shows a more stable scaling trend as the Tucker ranks increase from 8 to 64. The rank-32 configuration already achieves competitive performance, while the rank-64 configuration gives the strongest overall result for SRTA. This indicates that SRTA can increase adaptation capacity smoothly through its shared Tucker-decomposed representation space, without relying on a large bank of independent experts or an external gating network. Overall, the rank ablation demonstrates that SRTA provides a favorable balance between capacity, stability, and parameter efficiency across heterogeneous visual domains.

\begin{table*}[t]
\centering
\caption{Rank ablation across all PEFT and MoE-PEFT methods. Results are reported as mean $\pm$ standard deviation across multiple runs. For SRTA, $r_1$ and $r_2$ denote Tucker ranks, while $r_3$ is set to the number of domains.}
\label{tab:rank_ablation_all}
\resizebox{\textwidth}{!}{
\begin{tabular}{lccc ccccc}
\toprule
\textbf{Method} & \textbf{Rank-1} & \textbf{Rank-2} & \textbf{Extra Rank} 
& \textbf{PACS} & \textbf{VLCS} & \textbf{Office-Home} & \textbf{Digits-DG} & \textbf{NICO++} \\
\midrule

LoRA & 8  & -- & -- & $93.4{\scriptstyle \pm 0.6}$ & $81.1{\scriptstyle \pm 0.7}$ & $71.7{\scriptstyle \pm 1.2}$ & $91.6{\scriptstyle \pm 0.7}$ & $67.9{\scriptstyle \pm 1.4}$ \\
LoRA & 16 & -- & -- & $94.1{\scriptstyle \pm 0.5}$ & $81.5{\scriptstyle \pm 0.6}$ & $75.9{\scriptstyle \pm 1.4}$ & $92.2{\scriptstyle \pm 0.9}$ & $71.5{\scriptstyle \pm 0.8}$ \\
LoRA & 32 & -- & -- & $94.7{\scriptstyle \pm 0.4}$ & $82.4{\scriptstyle \pm 0.3}$ & $77.1{\scriptstyle \pm 0.6}$ & $92.9{\scriptstyle \pm 0.4}$ & $75.2{\scriptstyle \pm 0.9}$ \\
LoRA & 64 & -- & -- & $95.1{\scriptstyle \pm 0.1}$ & $82.8{\scriptstyle \pm 0.7}$ & $77.3{\scriptstyle \pm 0.3}$ & $93.8{\scriptstyle \pm 0.3}$ & $77.6{\scriptstyle \pm 0.3}$ \\
\midrule

DoRA & 8  & -- & -- & $93.5{\scriptstyle \pm 0.7}$ & $80.8{\scriptstyle \pm 1.1}$ & $72.1{\scriptstyle \pm 1.2}$ & $91.4{\scriptstyle \pm 0.7}$ & $67.1{\scriptstyle \pm 1.2}$ \\
DoRA & 16 & -- & -- & $93.9{\scriptstyle \pm 0.3}$ & $81.9{\scriptstyle \pm 0.8}$ & $76.1{\scriptstyle \pm 1.4}$ & $92.2{\scriptstyle \pm 0.5}$ & $71.7{\scriptstyle \pm 0.5}$ \\
DoRA & 32 & -- & -- & $94.8{\scriptstyle \pm 0.5}$ & $82.3{\scriptstyle \pm 0.6}$ & $77.3{\scriptstyle \pm 0.3}$ & $92.9{\scriptstyle \pm 0.3}$ & $74.8{\scriptstyle \pm 1.0}$ \\
DoRA & 64 & -- & -- & $95.0{\scriptstyle \pm 0.2}$ & $82.8{\scriptstyle \pm 0.3}$ & $77.3{\scriptstyle \pm 0.3}$ & $93.8{\scriptstyle \pm 0.4}$ & $77.4{\scriptstyle \pm 0.3}$ \\
\midrule

MALoRA & 8  & 4  & -- & $94.6{\scriptstyle \pm 0.6}$ & $83.8{\scriptstyle \pm 0.6}$ & $84.6{\scriptstyle \pm 0.3}$ & $88.7{\scriptstyle \pm 0.3}$ & $82.4{\scriptstyle \pm 0.7}$ \\
MALoRA & 16 & 8  & -- & $94.7{\scriptstyle \pm 0.6}$ & $83.3{\scriptstyle \pm 0.4}$ & $85.0{\scriptstyle \pm 1.0}$ & $90.4{\scriptstyle \pm 0.5}$ & $82.5{\scriptstyle \pm 0.2}$ \\
MALoRA & 32 & 16 & -- & $94.5{\scriptstyle \pm 0.5}$ & $84.4{\scriptstyle \pm 0.8}$ & $84.8{\scriptstyle \pm 0.8}$ & $89.2{\scriptstyle \pm 0.2}$ & $82.4{\scriptstyle \pm 0.2}$ \\
MALoRA & 64 & 32 & -- & $94.3{\scriptstyle \pm 0.5}$ & $82.8{\scriptstyle \pm 0.1}$ & $85.0{\scriptstyle \pm 0.0}$ & $90.2{\scriptstyle \pm 0.4}$ & $82.3{\scriptstyle \pm 0.3}$ \\
\midrule

MOELoRA & 8  & 4  & -- & $93.6{\scriptstyle \pm 0.5}$ & $84.1{\scriptstyle \pm 0.6}$ & $85.3{\scriptstyle \pm 0.1}$ & $88.6{\scriptstyle \pm 1.4}$ & $83.8{\scriptstyle \pm 0.9}$ \\
MOELoRA & 16 & 8  & -- & $94.0{\scriptstyle \pm 1.1}$ & $83.3{\scriptstyle \pm 1.9}$ & $84.7{\scriptstyle \pm 0.3}$ & $89.0{\scriptstyle \pm 1.7}$ & $83.5{\scriptstyle \pm 0.1}$ \\
MOELoRA & 32 & 16 & -- & $93.4{\scriptstyle \pm 1.7}$ & $84.6{\scriptstyle \pm 0.9}$ & $85.0{\scriptstyle \pm 0.4}$ & $88.9{\scriptstyle \pm 1.1}$ & $83.5{\scriptstyle \pm 1.2}$ \\
MOELoRA & 64 & 32 & -- & $94.5{\scriptstyle \pm 0.3}$ & $84.2{\scriptstyle \pm 0.5}$ & $85.2{\scriptstyle \pm 0.0}$ & $88.9{\scriptstyle \pm 1.3}$ & $83.3{\scriptstyle \pm 1.0}$ \\
\midrule

MoLoRA & 8  & -- & -- & $95.2{\scriptstyle \pm 0.2}$ & $84.7{\scriptstyle \pm 0.8}$ & $85.5{\scriptstyle \pm 0.4}$ & $91.3{\scriptstyle \pm 1.0}$ & $83.3{\scriptstyle \pm 0.3}$ \\
MoLoRA & 16 & -- & -- & $95.0{\scriptstyle \pm 0.2}$ & $84.1{\scriptstyle \pm 0.6}$ & $85.4{\scriptstyle \pm 0.3}$ & $92.2{\scriptstyle \pm 0.6}$ & $83.8{\scriptstyle \pm 0.5}$ \\
MoLoRA & 32 & -- & -- & $95.3{\scriptstyle \pm 0.3}$ & $85.4{\scriptstyle \pm 0.7}$ & $85.6{\scriptstyle \pm 0.5}$ & $92.4{\scriptstyle \pm 0.2}$ & $83.9{\scriptstyle \pm 0.5}$ \\
MoLoRA & 64 & -- & -- & $95.1{\scriptstyle \pm 0.3}$ & $84.4{\scriptstyle \pm 0.9}$ & $85.3{\scriptstyle \pm 0.6}$ & $91.7{\scriptstyle \pm 0.5}$ & $83.3{\scriptstyle \pm 0.3}$ \\
\midrule

SRTA & 8  & 8  & -- & $93.5{\scriptstyle \pm 0.6}$ & $84.4{\scriptstyle \pm 0.4}$ & $83.8{\scriptstyle \pm 0.4}$ & $88.7{\scriptstyle \pm 0.8}$ & $82.8{\scriptstyle \pm 0.3}$ \\
SRTA & 16 & 16 & -- & $94.6{\scriptstyle \pm 0.4}$ & $84.8{\scriptstyle \pm 0.3}$ & $84.2{\scriptstyle \pm 0.3}$ & $90.0{\scriptstyle \pm 0.3}$ & $82.7{\scriptstyle \pm 0.6}$ \\
SRTA & 32 & 32 & -- & $94.6{\scriptstyle \pm 0.5}$ & $84.3{\scriptstyle \pm 0.8}$ & $84.4{\scriptstyle \pm 0.5}$ & $91.7{\scriptstyle \pm 0.5}$ & $82.9{\scriptstyle \pm 0.8}$ \\
SRTA & 64 & 64 & -- & $95.2{\scriptstyle \pm 0.7}$ & $84.7{\scriptstyle \pm 0.3}$ & $84.8{\scriptstyle \pm 0.7}$ & $92.7{\scriptstyle \pm 0.6}$ & $83.0{\scriptstyle \pm 0.4}$ \\

\bottomrule
\end{tabular}
}
\end{table*}


\subsection{Routing Behavior Analysis}

\begin{figure}[H]
    \centering
    \includegraphics[width=\textwidth]{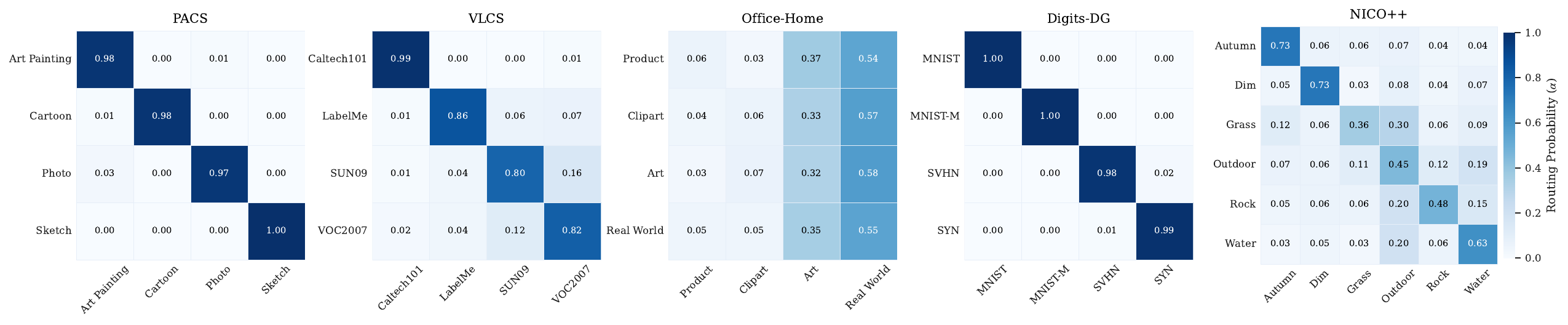}
    \caption{Intrinsic routing probabilities ($\alpha$) across core tensor slices.}
    \label{fig:routing_heatmaps}
\end{figure}

To better understand intrinsic routing, we visualize the average routing probabilities assigned to each Tucker core slice for every domain in Fig.~\ref{fig:routing_heatmaps}. The heatmaps show that SRTA learns domain-consistent routing patterns under auxiliary domain supervision. For visually distinct datasets such as PACS and Digits-DG, the routing distributions are sharp and nearly diagonal, indicating that each domain strongly activates a specific core slice. VLCS also shows largely domain-specific routing, with some sharing between related domains.

For more overlapping datasets such as Office-Home and NICO++, the routing distributions are softer, indicating greater sharing across domains. This supports the design of SRTA: visually separable domains can activate distinct pathways, while related domains can reuse shared Tucker core slices within a compact adaptation space. Since routing is supervised during training using domain labels, these visualizations should be interpreted as evidence of learned domain-aware routing rather than unsupervised domain discovery.

\section{Conclusion}

In this paper, we introduced \textbf{Self-Routed Tensor Adapters (SRTA)}, a compact parameter-efficient adaptation framework for heterogeneous multi-domain visual representation learning. SRTA removes the need for an external MoE router by deriving routing probabilities directly from the adapter's low-rank representation. These probabilities are used to blend slices of a shared Tucker core tensor, producing input-conditioned adaptation within a compact shared parameter space.

Experiments across five multi-domain visual classification benchmarks show that SRTA achieves competitive or slightly stronger average accuracy than MoE-style PEFT baselines while using substantially fewer trainable parameters. At rank 64, SRTA uses about $3.4\times$ fewer parameters than MoLoRA in the 4-domain setting and about $4.8\times$ fewer parameters in the 6-domain setting. These results suggest that SRTA's main strength is its accuracy-parameter trade-off: it provides MoE-level adaptation performance without relying on large independent expert banks or an external gating network.

The routing analysis further shows that SRTA learns interpretable domain-aware pathways, with sharper routing for visually distinct domains and softer sharing for overlapping domains. Overall, SRTA demonstrates that universal visual adaptation can be supported through a shared tensor-factorized adaptation space rather than through separate expert modules, making self-routed tensor adapters a promising building block for scalable adaptation of visual foundation models.

\section{Limitations}

While SRTA shows strong results on multi-domain image classification benchmarks, its application to broader visual tasks such as detection, segmentation, video understanding, and vision-language learning remains an important direction for future work. Our current experiments also use known domain labels for routing supervision, whereas real-world settings may involve unknown or continuously changing domains. Future extensions can explore unsupervised or adaptive routing mechanisms that discover domain structure automatically.

\section{LLM Usage}
The authors used an LLM to assist with grammatical and stylistic editing only. All changes were reviewed by the authors, who take full responsibility for the final manuscript.

\section{Acknowledgement}
The author acknowledges the GPU compute support provided by the Infosys Center for Artificial Intelligence at IIIT-Delhi. The author also thanks Siddharth Yadav from IIIT-Delhi for his support in creating the plots for this paper and Parth Goyal from IIIT-Delhi for his valuable review and suggestions that helped improve the writing of the paper.

%
%
\clearpage
\bibliographystyle{splncs04}
\bibliography{main}
\end{document}